\documentclass{article}
\usepackage{spconf,amsmath,graphicx}
\usepackage{xcolor}
\usepackage{cite}
\usepackage{booktabs}
\usepackage{multirow}
\usepackage{colortbl}
\usepackage{amsfonts}
\usepackage{comment}
\usepackage{float}
\usepackage{color}
\usepackage[table]{xcolor}
\usepackage[hidelinks]{hyperref}
\definecolor{headergrey}{gray}{0.85}
\definecolor{headermauve}{RGB}{223,214,235}
\definecolor{mauve}{RGB}{215,201,232}
\definecolor{headermauvedark}{RGB}{188,164,218}
\definecolor{headerlightgrey}{gray}{0.91}

\usepackage{enumitem}
\setlist{nosep, leftmargin=14pt}

\usepackage{mwe} 

\title{Refining Cytology Predictions with Conditional Random Fields}

\name{
\parbox{\linewidth}{\centering
Manon Dausort$^{\star}$, Tiffanie Godelaine$^{\star}$,  Karim El Khoury, Maxime Zanella, \\ Christophe De Vleeschouwer, and Beno\^it Macq\thanks{$^\star$The authors have contributed equally to this work.}}}

\address{ICTEAM, Université Catholique de Louvain, Belgium\\}

\begin{document}
%

\maketitle

\begin{abstract}
Vision-language models (VLMs) achieve strong zero-shot (ZS) classification on histology images but do not perform as well on cytology, whose stains and cell morphology differ markedly compared to histology. Conditional random fields (CRFs) can refine noisy VLM predictions by propagating information across patches, but existing CRF frameworks were designed for histopathology and do not transfer to cytology datasets, released as independent patch pools spanning multiple staining protocols. We introduce CytoCRF, which adapts the pairwise terms to cytology by targeting chromatin and cytology-specific staining, and further enrich the neighborhood of each potential term by combining multiple backbones. Across ten cytology datasets, CytoCRF outperforms existing CRF frameworks at every annotation budget, reaching +13.6 percentage points over the best baseline and +33.7 over ZS with only 50 annotations.Combining information from multiple backbones brings further gains, showing that the neighborhood topology matters more than the pairwise potential computed over it.
\end{abstract}

\begin{keywords}
Cytology Classification, Conditional Random Fields, Vision-Language Models
\end{keywords}

\section{Introduction}
\label{sec:intro}
Cytology involves examining under a microscope cells that have been collected through exfoliation or by aspiration from tissue, rather than tissue sections. It offers a minimally invasive alternative to biopsy and histology. Manual review, however, is labour-intensive and subject to inter-observer variability~\cite{darragh2013interrater}, motivating a large body of work on automating patch- or cell-level cytology classification~\cite{Jiang2023_review}. \\

Recently, histology-oriented vision-language models (VLMs), pretrained on large image-text corpora, have emerged as strong zero-shot (ZS) classifiers: models such as PLIP~\cite{huang2023plip}, CONCH~\cite{Lu2024conch} and Quilt~\cite{Oluchi2023quilt} produce patch-level predictions without any task-specific training. None of them was trained specifically for cytology, and their ZS predictions on cytology images remain noisy: histology VLMs have been shown to perform close to random on ZS cytomorphology tasks~\cite{kukuljan2026illusion}. Recent cytology-specific foundation models such as UniCAS~\cite{jiang2026unicas} close part of this gap through dedicated pretraining, but remain vision-only and offer no ZS classification by their own. Hence, a handful of expert annotations~\cite{song2022fsl,Pachetti2024} is required, e.g. to fine-tune the foundation model via low-rank adaptation~\cite{dausort2025finetuning}. Transductive methods further improves the predictions by leveraging the latent distribution of all samples, as Histo-TransCLIP~\cite{zanella2024boostingvlms}. We pursue this training-free, transductive direction and built on a probabilistic graph formulation. \\

Conditional random fields (CRFs) fit the transductive paradigm naturally as the probabilistic graph capture relations between patches. HistoCRF~\cite{godelaine2026histocrf} instantiates it for histology, introducing pairwise terms that both promote label diversity and propagate expert annotations to visually related patches. Building on this, SlideCRF~\cite{godelaine2026slidecrf} extends this framework to whole-slide images (WSIs), adding a spatial term that leverages the shared coordinates of patches on a single slide, together with biological cues based on texture and H\&E color. Neither method, however, has been applied to cytology. \\

In this work, we thus adapt the CRF-based paradigm to the cytology setting. In practice, no open-source WSI cytology dataset is publicly available; existing datasets are instead released as pools of independent cell images~\cite{Plissiti2018sipakmed,cai2024hicervix,Hussain2020}, with no information relating one patch to another. There is no shared spatial layout to exploit, and no single staining protocol to assume in advance: Papanicolaou, May-Grünwald-Giemsa and H\&E all co-occur across public cytology benchmarks. Hence, we introduce CytoCRF. It propose to drop the spatial term of SlideCRF and, adapt the biological cues to cytology by targeting the chromatin and a specific staining while revisiting how the neighborhood graph is built by leveraging information from multiple models. Across 10 cytology datasets, this improves the performance over the best baseline of \textbf{+13.6pp} balanced accuracy, even without any spatial term, and with only 50 annotations. This leads to an average increased performance of \textbf{+33.7pp} balanced accuracy. This performance is further (+2pp) increased when patches are connected using multiple models.




\section{Method}
\label{sec:method}
\subsection{Problem formulation}
\label{ssec:formulation}

We consider a set of patches $\mathcal{V}$, represented as a graph $\mathcal{G} = \{\mathcal{V}, \mathcal{E}\}$, where every patch $v \in \mathcal{V}$ is connected by a set $\mathcal{E}$ of edges to a neighborhood $\mathcal{N}_v \subset \mathcal{V}$. Each patch carries a hidden class label $y_v$ and three embeddings: (1) $\mathbf{f}_v$ from a VLM, (2) $\mathbf{g}_v$ from a second purely visual encoder, and (3) $\mathbf{f}_{k,v}$ representing the $k^{th}$ biological features extracted from the patch. These hidden variables are used to define the CRF objective function to minimize~\cite{godelaine2026histocrf,godelaine2026slidecrf}:
\begin{equation}
    f(\mathbf{y}) \;=\; \underbrace{\sum_{v \in \mathcal{V}} \phi_v(y_v)}_{\text{unary } \Phi_u}
    \;+\; \underbrace{\sum_{(v,w) \in \mathcal{E}} \Phi_{vw}(y_v, y_w)}_{\text{pairwise } \Phi_p},
    \label{eq:energy}
\end{equation}
where $\mathbf{y}$ is the label distribution. The unary potential penalizes assigning a label that is a poor match based on its own embedding, while the pairwise potential $\Phi_{vw}$ penalizes assigning an incompatible pair of labels to two connected patches, thereby encouraging consistency across patches.

The objective function in Eq.~\eqref{eq:energy} is minimized approximately via mean-field inference~\cite{Krahenbuhl2012}, which restricts the solution to a product of independent per-vertex marginals $Q(\mathbf{y}) = \prod_v Q_v(y_v)$. This yields an iterative update in which each marginal $Q_v$ is recomputed as the exponential of minus the unary potential $\phi_v$ combined with the pairwise potentials $\Phi_{vw}$, each averaged over the neighbor's current marginal $Q_w$ rather than its unknown true label. Repeating this update at every vertex in turn implements message passing along the graph $\mathcal{G}$: each patch's marginal is recomputed from its neighbors' current marginals, for a fixed number of propagation rounds.

\subsection{Unary potential}
\label{ssec:unary}

As in~\cite{godelaine2026histocrf}, the unary potential requires no task-specific training: a VLM encodes each patch into a visual embedding $\mathbf{f}_v$ and each class $l$ into a textual embedding $\mathbf{t}_l$ (averaged over a set of prompts). The unary potential is then defined based on the cosine similiarity:
\begin{equation}
    \phi_v(y_v = l) \;=\; -\log \operatorname{softmax}_l
        \!\left( \frac{\mathbf{f}_v \mathbf{t}_l^\top}{\lVert\mathbf{f}_v\rVert \lVert\mathbf{t}_l\rVert} \right).
    \label{eq:unary}
\end{equation}

\subsection{Pairwise potential}
\label{ssec:pairwise}

The pairwise potential is composed of three terms: \textit{diversity}, \textit{annotation} and \textit{biological}.

\vspace{2mm}
\noindent \textbf{Diversity and annotation terms.} We keep unchanged the two pairwise terms introduced by HistoCRF~\cite{godelaine2026histocrf}. 

The \textit{diversity} term connects each patch $v$ to its $|\mathcal{N}_v|$ most dissimilar patches, the similarity being defined by:
\begin{equation}
    \operatorname{sim} = \frac{\mathbf{g}_v \mathbf{g}_l^\top}{\lVert\mathbf{g}_v\rVert \lVert\mathbf{g}_l\rVert}.
    \label{eq:sim}
\end{equation}
This term penalizes assigning the same label to dissimilar patches, so predictions do not collapse onto a single class:
\begin{equation}
    \varphi_{vw}(y_v, y_w) \;=\; \delta_{y_v,y_w} \bigl(1 - \operatorname{sim}(\mathbf{g}_v,\mathbf{g}_w)\bigr).
    \label{eq:diversity}
\end{equation}

Let $\mathcal{A}$ be the set of annotated patches. The \textit{annotation} term connects each annotated patch $v \in \mathcal{A}$ to its $|\mathcal{M}_v|$ most similar patches. This term penalizes assigning them different labels, propagating expert supervision to visually related patches:
\begin{equation}
    \psi_{vw}(y_v, y_w) \;=\; (1-\delta_{y_v,y_w}) \operatorname{sim}(\mathbf{g}_v,\mathbf{g}_w).
    \label{eq:annotation}
\end{equation}

\noindent \textbf{Biological term.}
SlideCRF~\cite{godelaine2026slidecrf} extends this pairwise potential with biological terms, composed of \textit{texture} and \textit{color} cues. This term takes the same form as Eq.~\eqref{eq:annotation} to penalize patches with similar biological features $\textbf{f}_k$ to have different labels, the similarity being defined by a Gaussian kernel applied to the Euclidean distance between the features extracted from each patch: 
\begin{equation}
    \text{sim}_2(\mathbf{f}_{k, v}, \mathbf{f}_{k, w}) = \text{exp}
    \left(
    -\frac{||\mathbf{f}_{k, v} - \mathbf{f}_{k, w}||^2}{2\sigma^2}
    \right). 
    \label{eq:sim2}
\end{equation}
Unlike SlideCRF, we combine any subset of biological cue $\{f_1,\dots,f_K\}$ into a single pairwise term where each cue is divided by its own average magnitude $\operatorname{mean}(\xi_k)$ so a larger raw scale cannot dominate:
\begin{equation}
\begin{aligned}
        &\xi_{vw} = \frac{1}{K}\sum_{k=1}^K \frac{\xi_{k, vw}}{\operatorname{mean} (\xi_{k})}, \\
        &\xi_{k,vw}= (1-\delta_{y_v,y_w}) \operatorname{sim}_2(\mathbf{f}_{k,v},\mathbf{f}_{k,w}).
\end{aligned}
\label{eq:combined_bio}
\end{equation}
We choose May-Gr\"unwald-Giemsa staining and chromatin texture (cf. Section 3.1) for the final set of cues.
Another difference with SlideCRF is that each patch $v$ is connected to its $|\mathcal{M}_v|$ similar patches based on their embedding (Eq.~\eqref{eq:sim}). 

\vspace{2mm}
\noindent \textbf{Final pairwise potential.}
The three terms are then combined additively with weighting factors $\alpha$, $\beta$ and $\gamma$:
\begin{multline}
    \Phi_p \;=\; \alpha \Phi_{\text{div}} + \beta \Phi_{\text{ann}} + \gamma \Phi_{\text{bio}} 
    = \alpha \sum_{v}\sum_{w \in \mathcal{N}_v} \varphi_{vw} \\
    + \beta \sum_{v \in \mathcal{A}} \sum_{w \in \mathcal{M}_v} \psi_{vw} 
    + \gamma \sum_{v}\sum_{w \in \mathcal{M}_v} \xi_{vw}
\end{multline}

\section{Experiments}
\label{sec:exp} 
\subsection{Experimental setup}
\label{ssec:exp_setup}

\begin{figure}[!t]
    \centering
    \includegraphics[width=\linewidth]{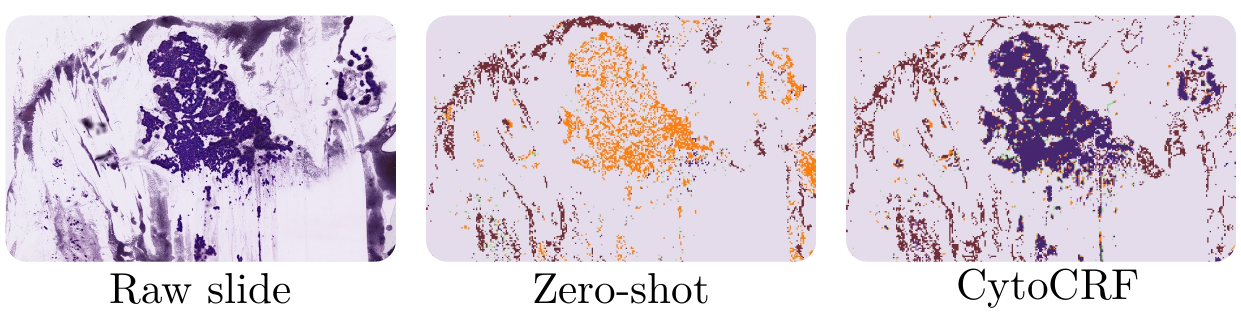}
    \caption{Qualitative illustration of CytoCRF on a thyroid fine-needle aspiration cytology slide. CytoCRF correctly identifies the \textcolor[HTML]{46266E}{proliferative-follicular} region, correcting the \textcolor[HTML]{ff7f0e}{colloid} misclassification of the KEEP zero-shot prediction, all while maintaining correct \textcolor[HTML]{722F37}{red blood cells} predictions.}
    \label{fig:wsi_illustration}
\end{figure}

\noindent \textbf{Datasets.} We evaluate on 10 cytology-oriented, patch-level classification datasets covering cervical (BMT~\cite{welch2024bmt}, APACC~\cite{kupas2024annotated}, SIPaKMeD~\cite{Plissiti2018sipakmed}, MLCC~\cite{Hussain2020}, HiCervix~\cite{cai2024hicervix}, Herlev~\cite{jantzen2005pap}), breast (FNAC~\cite{saikia2019comparative}), blood (BloodMNIST~\cite{yang2023medmnist}), body-cavity fluid (BCFC~\cite{sanyal2021machine}), and bone marrow cells (BMCD~\cite{shen2025large}). Unless stated otherwise, we report the balanced accuracy of the final mean-field inference, averaged over 10 seeds per dataset and then over the 10 datasets.

\noindent \textbf{Foundation models.} As unary $\mathbf{f}_v$ candidates, PLIP~\cite{huang2023plip}, CONCH~\cite{Lu2024conch}, Quilt-B16~\cite{Oluchi2023quilt} and KEEP~\cite{ZHOU2026777}, with eight vision-only encoders as pairwise $\mathbf{g}_v$ candidates, including the visual encoder of the four VLMs, GigaPath~\cite{Xu2024gigapath}, UNI~\cite{Chen2024uni}, DinoBloom~\cite{koch2024dinobloom} and UniCAS~\cite{jiang2026unicas}. 

\vspace{1mm}

\noindent \textbf{Biological cues.} We test two color and three texture cues, one from SlideCRF's own implementation. The feature \texttt{he}~\cite{ruifrok2001quantification} deconvolves the patch into the hematoxylin eosin space and concatenates the mean and standard deviation of each stain channel; \texttt{mgg} does the same with a May-Gr\"unwald-Giemsa matrix, better suited to cytology than the histological \texttt{he} stain. \texttt{texture}~\cite{haralick1973textural,ojala2002multiresolution} concatenates gray-level co-occurrence statistics with a local binary pattern histogram of the grayscale patch; \texttt{chrom} applies the same statistics to the nucleus-stain channel; \texttt{density} thresholds (Otsu) that channel into a nucleus mask and computes its area fraction, a nucleus-to-cytoplasm ratio proxy, and its spatial variance.

\vspace{1mm}

\noindent \textbf{Annotation strategy.} \textit{random} uniformly samples patches; \textit{error} selects patches where the model's current prediction disagrees with the ground-truth label, simulating a pathologist correcting the model's mistakes; and \textit{oscillation} selects patches whose predicted label flips most often across the mean-field propagation steps of the previous round.

\subsection{CytoCRF outperforms existing CRF frameworks}
Table~\ref{tab:main_results_bis} reports balanced accuracy averaged over our ten cytology datasets for our base configuration: we combine the color cue \texttt{mgg} with the texture cue \texttt{chrom}, and use KEEP for the unary potential and DinoBloom for the pairwise potential. We compare it against ZS predictions and two CRF baselines: HistoCRF-style (diversity and annotation terms only) and SlideCRF-style (addition of the \texttt{texture} and \texttt{he} cues, with a \emph{dissimilar} graph). ZS classification with KEEP VLM reaches only 26.7\% balanced accuracy. Our configuration leads at every budget (41.2--60.0\%), and its margin over HistoCRF-style grows rather than shrinks as more annotations are added (+5.8pp at budget 5 to +13.6pp at budget 50), showing that the gain from our biological pairwise term compounds with more supervision instead of being absorbed by it. SlideCRF-style, run on its own historical dissimilar graph, only overtakes ZS from budget 10 onward and remains the weakest CRF configuration throughout. Figure~\ref{fig:wsi_illustration} illustrates CytoCRF qualitatively on a real WSI image.

\subsection{Ablation experiments}
\label{ssec:FM_ablation}
\noindent \textbf{Unary and pairwise potentials.} Table~\ref{tab:FM_grid} reports the full grid crossing the four unary VLM candidates $\mathbf{f}_v$ and the eight pairwise visual encoder candidates $\mathbf{g}_v$, each evaluated with our base configuration (\texttt{mgg} and \texttt{chrom}). The pairwise encoder matters far more than the unary VLM: DinoBloom is the best pairwise choice for every unary candidate and Quilt-B16 the worst, a gap of over 14pp regardless of the unary VLM. The four vision-only encoders (UNI, GigaPath, DinoBloom, UniCAS) also consistently outperform the four VLMs' own vision encoder as pairwise candidates, confirming that a dedicated visual encoder captures finer-grained patch similarity than a VLM's vision encoder. KEEP paired with DinoBloom reaches the best overall balanced accuracy. \\

\begin{table}[!t]
\centering
\caption{Mean balanced accuracy of the different pipelines (with $\alpha=0.1$, $\beta=0.5$, $\gamma=0.5$ considering the \textit{random} strategy) across annotation budgets of $\{5,10,25,50\}$. $\pm$ is the standard deviation over 10 seeds, averaged across the 10 datasets.}
\label{tab:main_results_bis}
\resizebox{\linewidth}{!}{
\begin{tabular}{l||cccl}
    \toprule
    Method & 5 & 10 & 25 & 50 \\
    \hline
    ZS & 26.7 & 26.7 & 26.7 & 26.7 \\
    HistoCRF~\cite{godelaine2026histocrf} & 35.4$\pm$1.9 & 38.1$\pm$2.6 & 42.4$\pm$2.8 & 46.4$\pm$2.0 \\
    SlideCRF~\cite{godelaine2026slidecrf} & 25.7$\pm$0.9 & 27.2$\pm$1.0 & 29.2$\pm$1.3 & 32.7$\pm$1.8 \\
     \rowcolor{mauve} \textbf{CytoCRF} (Ours) & \textbf{41.2$\pm$5.3} & \textbf{47.4$\pm$4.9} & \textbf{55.3$\pm$3.8} & \textbf{60.0$\pm$2.3} \\
    \bottomrule
\end{tabular}}
\end{table}

\begin{table}[!t]
\centering
\caption{Mean balanced accuracy of every unary / pairwise foundation model pair (with \texttt{mgg} and \texttt{chrom}, considering the \textit{random} annotation strategy with 25 annotations budget. }
\label{tab:FM_grid}
\resizebox{\linewidth}{!}{
    \begin{tabular}{l||cc>{\columncolor{headermauve}}cc}
    \toprule
    Pairwise $\backslash$ Unary & CONCH & Quilt-B16 & KEEP & PLIP \\
    \hline
    CONCH & 42.02 & 44.04 & 44.27 & 44.32 \\
    Quilt-B16 & 38.67 & 40.37 & 40.03 & 40.73 \\
    KEEP & 42.61 & 45.78 & 46.27 & 46.64 \\
    PLIP & 39.78 & 41.87 & 40.94 & 42.01 \\
    UNI & 44.44 & 46.64 & 46.28 & 46.53 \\
    GigaPath & 50.39 & 52.83 & 52.52 & 53.50 \\
    \rowcolor{headermauve} DinoBloom & 52.59 & 54.09 & \cellcolor{headermauvedark}\textbf{54.71} & 54.40 \\
    UniCAS & 47.32 & 49.25 & 48.96 & 49.51 \\
    \bottomrule
    \end{tabular}
    }
\end{table}

\begin{table}[!t]
\centering
\caption{Balanced accuracy \% for every biological cue, solo (diagonal) and pairwise (off-diagonal), on KEEP and DinoBloom pair and considering \textit{random} strategy with 25 annotations budget.}
\label{tab:cue_ablation}
\resizebox{\linewidth}{!}{%
    \begin{tabular}{l||cccc>{\columncolor{headermauve}}cc}
    \toprule
     & \texttt{he} & \texttt{mgg} & \texttt{texture} & \texttt{density} & \texttt{chrom} \\
    \hline
    \texttt{he} & 54.01 & 54.89 & 54.82 & 54.42 & 54.96 \\
    \rowcolor{headermauve} \texttt{mgg} &  & 54.71 & 54.98 & 54.51 & \cellcolor{headermauvedark}\textbf{55.28} \\
    \texttt{texture} &  &  & 54.31 & 54.36 & 54.83 \\
    \texttt{density} &  &  &  & 53.76 & 54.60 \\
    \texttt{chrom} &  &  &  &  & 54.41 \\
    \bottomrule
    \end{tabular}
}
\end{table}

\begin{table}[!t]
\centering
\caption{Pairwise graph ensemble (balanced accuracy \%, \texttt{mgg}+\texttt{chrom}, budget 25).}
\label{tab:graph_ensemble}
\resizebox{0.65\linewidth}{!}{
    \begin{tabular}{l||c}
    \toprule
    Additional FM & Bal. Acc. (\%) \\
    \hline
    DinoBloom & 55.28 \\
     \quad + KEEP & 56.63 \\
     \quad + KEEP, GigaPath & \textbf{58.38} \\
    \bottomrule
    \end{tabular}}
\end{table}

\noindent \textbf{Biological cues.} Table~\ref{tab:cue_ablation} reports all five biological cues, individually (diagonal) and in every pairwise combinations (off-diagonal), combined via Eq.~\eqref{eq:combined_bio} at equal weight over the \emph{similar} neighborhood graph.  \texttt{chrom} is the strongest solo cue and \texttt{mgg}+\texttt{chrom} the strongest pair; the five solo cues are otherwise close. We also ablate the neighborhood graph itself: switching SlideCRF's dissimilar to the similar graph improves balanced accuracy by 20pp (34.6--54.7\%). Neighbors from $\mathbf{g}_v$ are thus more informative than either alternative graph. This holds beyond solo cues: our default pair \texttt{mgg}+\texttt{chrom} reaches 55.3\% on similar versus 43.9\%/43.8\% on \texttt{mgg}'s or \texttt{chrom}'s own raw-feature graph, and \texttt{chrom} alone reaches 54.4\% versus 36.2\% on dissimilar. 

\vspace{1mm}

\noindent \textbf{Pairwise graph ensemble.} Beyond the single KEEP and DinoBloom graph, we test enlarging the set of dissimilar $\mathcal{N}_v$ and similar $\mathcal{M}_v$ patches with the neighbors found in one or two additional FMs' own embedding spaces. The potential itself is still computed from $\mathbf{g}_v$ (DinoBloom): only the set of candidate neighbors grows, increasing the relations between patches. Each additional FM brings a further gain, from +1.35pp for KEEP to +3.10pp for KEEP and GigaPath, with no saturation across the three configurations tested. 

\vspace{1mm}

\begin{figure}[!t]
    \centering
    \includegraphics[width=\linewidth]{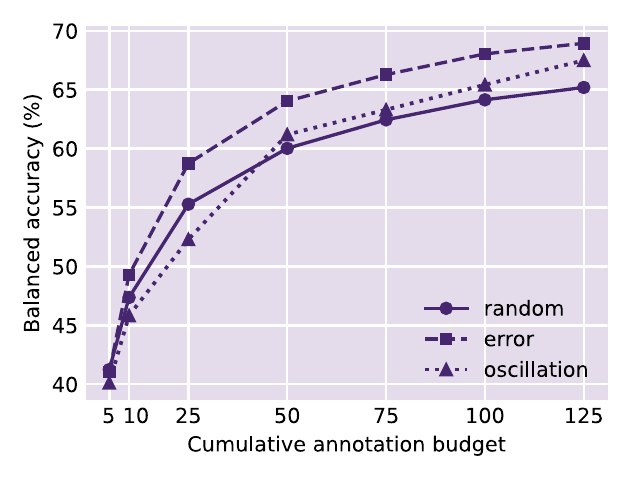}
    \vspace{-10mm}
    \caption{Balanced accuracy vs.\ cumulative annotation budget for the three annotation strategies (\texttt{mgg}+\texttt{chrom}, KEEP+DinoBloom).}
    \label{fig:annotation_strategy}
\end{figure}

\noindent \textbf{Annotation strategy.} Finally, we compare our three annotation strategies (Sec.~\ref{ssec:exp_setup}) for our default configuration (\texttt{mgg}+\texttt{chrom} on KEEP + DinoBloom), over $R \in \{1,2,5,10,15,20,25\}$ human-in-the-loop rounds of $5$ annotations each (Figure~\ref{fig:annotation_strategy}). Correcting the model's wrong patches (\textit{error}) is an oracle upper bound on what a human-in-the-loop protocol could gain as it relies on the ground truth label. At the smallest budget (5 annotations, a single round with no propagation history yet to exploit), the three strategies are within 1.2pp of each other. From 50 to 100 annotations, \textit{random} and \textit{oscillation} have performances close to each other. The gains from \textit{oscillation} do not increase until 125 annotations, suggesting that its signal requires more propagation rounds to become reliable.


\section{Conclusion}

\label{sec:conclu}
We adapted a CRF-based refinement of vision-language model predictions to cytology, replacing the spatial neighborhood assumption of prior whole-slide work with a systematic search over FM pairs and biological/color cues suited to independent cytology patch collections. On ten cytology datasets, CytoCRF beats both existing CRF-based baselines at every annotation budget we tested, with the margin widening rather than narrowing as more annotations become available. Combining the pairwise neighborhood across additional FMs brings a further, uncorrelated gain that does not saturate within the range we tested. Together these results show that domain-specific care in choosing the graph, not just the pairwise term, is what makes CRF refinement work on cytology's patch-independent, multi-stain benchmarks. In every ablation we ran, the neighborhood topology itself, whether set by the pairwise FM, the similar/dissimilar graph, or the multi-FM ensemble, mattered more than any change to the potentials computed over it. A natural direction for future work is therefore to study this topology directly, for instance by learning or adapting the neighborhood structure rather than fixing it to a handful of hand-picked FMs and similarity thresholds.

\section{Compliance with Ethical Standards}

This is a study for which no ethical approval was required.

\section{Acknowledgments}

\label{sec:acknowledgments}
M. Dausort and T. Godelaine are funded by the MedReSyst project, supported by FEDER and the Walloon Region.
Computational resources were made available on the Lucia infrastructure (Walloon Region grant n°1910247). 

\bibliographystyle{IEEEbib}

\small\bibliography{refs}

\end{document}